\documentclass[10pt,twocolumn,letterpaper]{article}

\usepackage{cvpr}               
\usepackage{graphicx}
\usepackage[table]{xcolor}
\usepackage{placeins}

\definecolor{cvprblue}{rgb}{0.21,0.49,0.74}
\usepackage[pagebackref,breaklinks,colorlinks,allcolors=cvprblue]{hyperref}

\def\paperID{}                 
\def\confName{}                
\def\confYear{}                

\title{LiDARDraft: Generating LiDAR Point Cloud from Versatile Inputs}

\author{
Haiyun Wei \quad Fan Lu \quad Yunwei Zhu \quad Zehan Zheng \\
Weiyi Xue \quad Lin Shao \quad Xudong Zhang \quad Ya Wu \quad Guang Chen$^{*}$\\
\\
TongjiUniversity \\
{\tt\small \{2311399, lufan, 2432040, zhengzehan, xwy, 2332976,}
\\
{\tt\small east\}@tongji.edu.cn, qpo144@163.com, guangchen@tongji.edu.cn}
\\}

\begin{document}
\maketitle


\begin{abstract}
Generating realistic and diverse LiDAR point clouds is crucial for autonomous driving simulation. Although previous methods achieve LiDAR point cloud generation from user inputs, they struggle to attain high-quality results while enabling versatile controllability, due to the imbalance between the complex distribution of LiDAR point clouds and the simple control signals. To address the limitation, we propose LiDARDraft, which utilizes the 3D layout to build a bridge between versatile conditional signals and LiDAR point clouds. The 3D layout can be trivially generated from various user inputs such as textual descriptions and images. Specifically, we represent text, images, and point clouds as unified 3D layouts, which are further transformed into semantic and depth control signals. Then, we employ a rangemap-based ControlNet to guide LiDAR point cloud generation. This pixel-level alignment approach demonstrates excellent performance in controllable LiDAR point clouds generation, enabling``\textit{simulation from scratch}", allowing self-driving environments to be created from arbitrary textual descriptions, images and sketches.
\end{abstract}    
\section{Introduction}
\label{sec:intro}
Autonomous vehicles require extensive data for simulation, training and validation. However, collecting real-world physical data is costly, unsafe, and difficult to scale, significantly hindering advancements in the field of autonomous driving. The emergence of LiDAR point cloud generation offers a solution by leveraging deep generative models to learn and resample LiDAR point cloud distributions. In particular, recent advancements in conditional scene generation, guided by text, images, and semantic cues, facilitate the generation of point clouds that meet specific requirements, potentially enabling ``\textit{simulation from scratch}". This formulation allows for the conditional creation of self-driving simulation scenes based on a single sentence, a casual image, or even a rough sketch.

Early methods~\cite{Deep,ultralidar} have attempted to employ Variational AutoEncoders (VAEs)~\cite{VAE} for LiDAR point cloud generation, and Li et al.~\cite{Deep,projectedgan} have also employed Generative Adversarial Networks (GANs)~\cite{GAN} to it. But they struggled to produce high-quality results. Recently, Diffusion Models\cite{DDPMS,Score-Based,latent} have demonstrated significant progress across various content generation tasks, such as image generation~\cite{latent,podell2023sdxl,xiao2024omnigen}, video generation~\cite{blattmann2023stable,ho2022imagen}, and 3d object generation~\cite{wang2024prolificdreamer,yi2024gaussiandreamer,vahdat2022lion,lyu2021conditional}. Moreover, several studies have explored their application to generating LiDAR point clouds~\cite{lidargen,lidardiffusion,lidardm}. For instance, LiDARGen~\cite{lidargen} first performs Diffusion Models on the projected range image of LiDAR point clouds. While LiDARGen~\cite{lidargen} has improved quality, they still face challenges in generating LiDAR point clouds conditioned on versatile user inputs. An innovative approach LiDARDM\cite{lidardm} uses layouts to guide scene generation. However, since the point clouds are projected via meshes rather than generated directly, authenticity remains a concern. Methods like LiDARDiffusion~\cite{lidardiffusion} have sought to address this by encoding user input(text/image) as global features, integrating them into the generation process of controlled outputs. However, the vastness and information-rich nature of automatic driving scenes make it challenging to capture features adequately with a simple input, resulting in weak correspondences that hinder conditional generation. This raises the question: Can we devise an approach that generates LiDAR point clouds from simple inputs, thereby enabling ``\textit{simulation from scratch}"?



{ Based on the above considerations, we propose to use 3D layouts as a unified conditional representation to bridge the gap between diverse user inputs and LiDAR point clouds. A 3D layout consists of a set of semantic primitives with simple geometry (e.g., cuboids, ellipsoids, etc.). This approach offers two key advantages: \textit{(1) Ease of Creation}: Due to the simplicity of the geometry of semantic primitives, users can readily create 3D layouts using 3D modeling software such as Blender\cite{djurayev2023understanding} or AutoCAD\cite{yarwood2010introduction}. In addition, translating various inputs into 3D layouts is straightforward. \textit{(2) Detailed Control}: 3D layouts encapsulate both the basic geometric structure and semantic information of 3D scenes, providing dense conditional signals for the generation process and enabling precise semantic and geometric control over the generated results.}

{ Built upon the 3D layout representation, we propose LiDARDraft, a layout-guided LiDAR point cloud generation framework. Specifically, we depict text, image, and point cloud as layout, which are subsequently projected into range image through the method of raycasting and subsequently utilized as input for ControlNet~\cite{controlnet} to control the process of unconditional LiDAR point cloud generation.
}


In summary, our contributions are as follows:

\textbf{A Unified Conditional Framework for LiDAR Point Cloud Generation:} We introduce a novel, unified framework that supports multi-modal conditional generation for LiDAR point clouds, supporting conditions including text, image, and point cloud. This is, to our knowledge, the most comprehensive and transferable framework in conditional point cloud generation, enabling seamless adaptation to diffusion-based models through fine-tuning.

\textbf{Direct Layout-to-Point Cloud Control:} For the first time, we explore direct layout-to-point cloud generation with point-wise control, achieving an optimal balance between controllability and diversity. This approach advances control granularity within generative models for LiDAR point clouds.



\section{Related Work}
\label{sec:relatedwork}
\textbf{LiDAR Simulation} plays a vital role in robotics and autonomous vehicles. Conventional methods, such as CARLA~\cite{carla} and AirSim~\cite{airsim}, employed raycasting-based physical techniques. Simulators first established a 3D simulated environment that included both static and dynamic objects, within which virtual sensors were positioned. These sensors cast rays that intersected with objects. Depth and orientation were calculated through intersection points. These methodologies were straightforward and easy to implement, making them popular in robotic simulations~\cite{gzscenic,robot}. However, physical-based simulation methods were constrained due to the high demand for 3D assets and the difficulties in bridging the simulation-to-reality gap regarding asset design. Recent studies~\cite{manivasagam2020lidarsim,wang2022learning,amini2022vista} sought to overcome these limitations with data-driven techniques. LiDARSim~\cite{manivasagam2020lidarsim} employed gathered LiDAR sequences to reconstruct maps and dynamic entities. Simulation-to-reality discrepancies like ray dropping~\cite{lidardm,nakashima2023generative} and snow~\cite{hahner2022lidarsnow} were also addressed. Methods like~\cite{lidarnerf,huang2023neural,zheng2024lidar4d,xue2024geonlf} enhanced asset reconstruction through approaches based on NeRF. LiDAR4D~\cite{zheng2024lidar4d} incorporated dynamic NeRF to effectively capture the dynamic characteristics inherent in LiDAR point clouds, while GeoNLF~\cite{xue2024geonlf} presented a pose-free method for registration during reconstruction. However, reconstructing scenes from sensor data required LiDAR scan frames, which entailed substantial costs.

\noindent\textbf{LiDAR Generation} was primarily employed in the context of autonomous driving simulations, significantly expanding the range of training and testing data sources. The predominant methods~\cite{lidargen, Deep, Synthesis} projected LiDAR point clouds into range images and applied Diffusion Models on the resulting 2D range representations, while others~\cite{lidardiffusion, lidardm, ultralidar} mapped these projected range images into a latent space. Among these approaches, Diffusion Models emerged as the most widely adopted technique. LiDARVAE~\cite{Deep} introduced the concept of deep generative modeling to LiDAR point clouds, exploiting advancements in both variational autoencoders (VAE)\cite{VAE} and generative adversarial networks (GAN)\cite{GAN} to construct a generated scenario of point clouds. LiDARGen~\cite{lidargen} was the first to apply Diffusion Models to LiDAR point cloud generation, whereas LiDARDiffusion~\cite{lidardiffusion, lidardm} introduced multimodal conditions to enhance control over the generative process. Additionally, UltraLidar~\cite{ultralidar} incorporated VQ-VAE~\cite{vqvae} to encode 3D point clouds into a discrete codebook. Due to the vector-quantized method, it could be conveniently applied for scene editing and could remove or insert actors from or into the generated scene with ease.

\noindent\textbf{Conditional Generation} was developed to incorporate conditions into generative models to sample results that aligned with expectations. Classifier-based guidance~\cite{sdedit,lidargen,generativemodelingestimate} used the gradients of a classifier to influence the generative process. The generative model could be guided toward regions of the data distribution, enabling Diffusion Models to generate outputs conditioned on specific classes. Diverging from classifier-based guidance, classifier-free guidance~\cite{classifier} utilized an implicit classifier as a substitute for the explicit classifier, allowing the gradients of the classifier to be expressed in terms of generative probabilities. ControlNet~\cite{controlnet} introduced a method for enhancing the controllability of a pre-trained Diffusion Model by adding a designed hypernetwork and fine-tuning it. It could simply integrate single or multiple conditions into the original Diffusion Model. The layout conditions were successfully added to the Diffusion Models using ControlNet~\cite{controlnet}, generating point clouds that aligned with the specified layout.

\section{Methodology}
 \vspace{-2pt}
\begin{figure*}[htbp]
\centering
\includegraphics[width=1\textwidth]{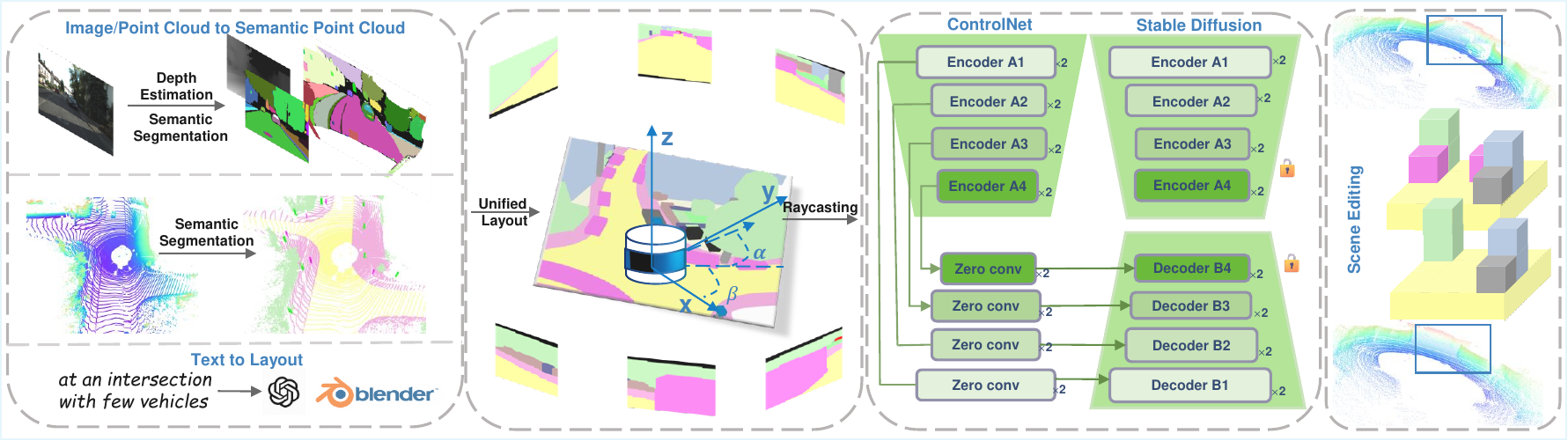}
\caption{\textbf{Overview of LiDARDraft.} Various inputs are unified into layout representations and projected into range images using RayCasting, which are then fed into ControlNet to guide unconditional LiDAR point cloud generation. \textbf{Image} first undergoes semantic segmentation and depth estimation to form a pseudo-point cloud and then clustered to create the layout. \textbf{Point cloud} is semantically segmented and then clustered. \textbf{Text} can generate layout using a Large Language Model. Moreover, we can easily modify the layout to edit the scene.}
\label{fig:pipeline}
\end{figure*}
\vspace{-2pt}

\subsection{LiDAR Point Cloud Representation}
 LiDAR point clouds can be represented with multiple representations, including 3D point sets, voxel grids, and range images. In this work, we adopt the range image as the basic representation for LiDAR point clouds due to its effective 2D parameterization of LiDAR point clouds. Specifically, each pixel in a range image corresponds to a unique direction in the LiDAR sensor's field of view, where the pixel value denotes the distance (or range) to the closest object along that direction. This representation maintains spatial structure and is  well-suited for convolution-based generative networks~\cite{rangenet,efficient}. Formally, a range image can be denoted as $x \in \mathbb{R}^{H \times W}$, where $H$ and $W$ correspond to the vertical and horizontal angular resolutions of the LiDAR sensor. For each pixel defined by its normalized 2D coordinates and range value, we can directly compute depth, yaw and pitch based on predefined sensor parameters.


For each pixel in the range image, defined by its normalized 2D location and range value, we can directly compute its depth, yaw and pitch given predefined sensor parameters. 
Specifically, for a pixel with depth \( \text{depth} \), we calculate its 3D coordinates \( p = [\alpha, \beta, \gamma] \) using the following equations:
\begin{equation}
    \alpha = \cos(\text{yaw}) \times \cos(\text{pitch}) \times \text{depth},
\end{equation}
\begin{equation}
    \beta = -\sin(\text{yaw}) \times \cos(\text{pitch}) \times \text{depth},
\end{equation}
\begin{equation}
    \gamma = \sin(\text{pitch}) \times \text{depth},
\end{equation}
where \(\text{yaw}\) and \(\text{pitch}\) represent the horizontal and vertical angles relative to the LiDAR sensor’s origin. These angles are derived from the pixel’s location in the range image and the sensor’s angular resolution, translating each pixel’s 2D position and depth into precise 3D coordinates. This structured representation enables efficient integration of point cloud data into our layout-conditioned Diffusion Model, enhancing the generation’s spatial fidelity and realism for autonomous driving simulations.

\subsection{LiDARDraft}
 LiDARDraft is a diffusion-based conditional LiDAR point cloud generation method designed to produce realistic LiDAR data from coarse 3D layout representations. Specifically, we first transform the layout into 2D control signals to align with the range image representation. Next, an unconditional LiDAR point cloud generation model is trained to capture the basic geometric distribution of LiDAR point clouds. Finally, the transformed control signals are incorporated into a conditional generation pipeline.

\noindent \textbf{Layout as a unified conditional representation.} 
 A scene layout consists of a set of semantic primitives, each represented by a triangular mesh with a semantic label and a simple geometric structure. For example, cars can be represented by cuboids, vegetation by ellipsoids, and roads by planes. By combining these primitives with appropriate spatial relationships, the layout encapsulates a coarse semantic and geometric distribution of the scene, providing precise control signals to guide the generative model. Thanks to the simplicity and flexibility of semantic primitives, drafting a desired scene in 3D modeling software like Blender is straightforward. Additionally, retrieving layouts from various data sources is simple.LiDARDraft unifies diverse user inputs into a common layout representation. Specifically, text inputs are processed by a large language model (e.g., GPT-5) to generate layout code, which is then used to construct a scene in Blender. For image inputs, depth estimation and semantic segmentation models are applied; the outputs are clustered based on semantic categories to obtain bounding boxes, which are used to build the layout scene. Similarly, point cloud inputs are fed into a semantic segmentation model, and the points are clustered by semantic class into bounding boxes to construct the corresponding layout.

\noindent \textbf{Scene raycasting.} 
 After constructing the layout of a scene, the key challenge is transforming the layout into effective conditional signals. As previously mentioned, we use range images as the primary representation for LiDAR point clouds. To provide pixel-aligned conditional signals, we convert layouts into a range image-like format that preserves the scene's semantic and geometric structure. Specifically, we employ raycasting to sample the conditional range image, with the LiDAR's position as the origin, emitting rays outward. Each ray corresponds to a pixel in the range image, and the number and direction of the rays are determined by the LiDAR sensor's field of view and resolution. The raycasting operation identifies the intersection point between each ray and a mesh triangle, allowing us to compute depth from the intersection coordinates and extract semantic information from the intersected triangle ID. As a result, the conditional range image is the concatenation of the semantic range image and depth range image.

\noindent \textbf{Layout-guided Generation}
Given the pixel-aligned conditional signals, a straightforward approach to achieve conditional generation is to directly concatenate the conditional signals with the denoising latents as a unified input to denoising networks. However, we observe that this paradigm fails to provide satisfactory and flexible controllability. Thus, we start by training a Diffusion Model for unconditional LiDAR point cloud generation. Thereafter, based on the pretrained weights, we integrate conditional signals by finetuning a ControlNet.

\noindent\textbf{Unconditional generation.} Diffusion Models~\cite{DDPMS,Score-Based, latent} have recently achieved remarkable success in generative tasks, leveraging iterative denoising to generate high-quality, high-fidelity data samples from latent distributions. Originally applied in image generation, Diffusion Models have been adapted to LiDAR point cloud generation, capturing fine-grained details and realistic object structures through gradual noise-to-signal refinement. State-of-the-art methods like LiDARGen~\cite{lidargen} and LiDARDiffusion~\cite{lidardiffusion} have demonstrated that Diffusion Models are particularly well-suited for point cloud generation, due to their ability to model complex data distributions with high fidelity~\cite{lidargen}. In our approach, we employ a Diffusion Model to generate point clouds conditioned on layouts, where the model iteratively refines initial noisy layouts into realistic point cloud scenes. This allows for precise control over the generated structure of LiDAR point clouds, bridging the gap between spatial layouts and 3D point clouds.

\noindent\textbf{Conditional generation.} To integrate conditional control into the diffusion-based generation process, We utilize ControlNet~\cite{controlnet}, a model architecture designed to control generative networks through additional conditional inputs without retraining the original model. ControlNet~\cite{controlnet} functions by introducing a set of condition-driven layers that guide the generation process in alignment with specific input requirements, such as layout configurations. With ControlNet~\cite{controlnet}, we achieve fine-grained control over LiDAR point cloud generation, enabling various conditioning scenarios by incorporating information directly from layouts. Importantly, ControlNet~\cite{controlnet} enables flexibility in mixing conditions, where layouts can be modified dynamically to influence the generated scene’s content and structure, a critical feature for diverse autonomous driving simulation requirements. By leveraging ControlNet’s~\cite{controlnet} ability to impose structure, we avoid full retraining of the underlying Diffusion Model, significantly enhancing both computational efficiency and adaptability for multiple control conditions. For conditional generation, we utilize the ControlNet~\cite{controlnet} framework:as illustrated in \autoref{fig:pipeline}, our ControlNet consists of four residual blocks with channel dimensions of \{128, 128, 256, 256\}. Meanwhile, the encoder of the Stable Diffusion model is composed of four residual blocks with the same channel configuration, i.e., \{128, 128, 256, 256\}. The decoder comprises four refinement blocks, whose channel dimensions are \{256, 256, 128, 128\}, respectively. Each block is repeated twice. The output of each block of controlnet is passed through a zero convolution layer and then connected to the corresponding decoder block in Stable Diffusion. We freeze the parameters of the unconditional generation model and fine-tune only the ControlNet. With this setup, our model achieves visually compelling conditional generation results in just 5,000 training steps, reducing the training cost by 95,000 steps compared to training from scratch. To further accelerate convergence, we initialize the parameters of ControlNet by copying those from the pretrained Stable Diffusion model.

\subsection{Training Objectives}
Our training objective consists of two parts: one for unconditional generation, and the other for conditional generation.
The unconditional generative model is based on a denoising score network \( S_\theta \). In our experiments, we found that noise adjustment plays a critical role in influencing sampling outcomes. We extend the score network \( S_\theta(x, \sigma_i) \), to be dependent on the current noise perturbation level \( \sigma_i \). During training, we follow the noise-conditioned score matching model as described in \cite{generativemodelingestimate}. A multi-scale loss function is employed, where each noise level is assigned a weight. The loss function is defined as:

\begin{footnotesize}
\begin{equation}
L = \frac{1}{2L} \sum_{i=1}^{L} \sigma_i^2 \, \mathbb{E}_{p_{\text{data}}(x)} \, \mathbb{E}_{\tilde{x} \sim \mathcal{N}(x, \sigma_i^2 I)} \left[ \left\| S_\theta(\tilde{x}, \sigma_i) + \frac{\tilde{x} - x}{\sigma_i^2} \right\| \right]
\end{equation}
\end{footnotesize}
where \( \tilde{x} \) represents the randomly perturbed noisy signal at each noise level, and \( \sigma_i \) is the standard deviation of the noise distribution.
 Given an input condition \( x_c \), the loss function for conditional generation is defined as:

\begin{equation}
\scalebox{0.9}{$
L = \frac{1}{2L} \sum_{i=1}^{L} \sigma_i^2 \, \mathbb{E}_{p_{\text{data}}(x)} \, \mathbb{E}_{\tilde{x} \sim \mathcal{N}(x, \sigma_i^2 I)} \left[ \left\| S_\theta(\tilde{x}, x_c, \sigma_i) + \frac{\tilde{x} - x}{\sigma_i^2} \right\| \right]
$}
\end{equation}


\section{Experiments}
\subsection{Setup}
\begin{figure}
\centering
\includegraphics[width=0.5\textwidth]{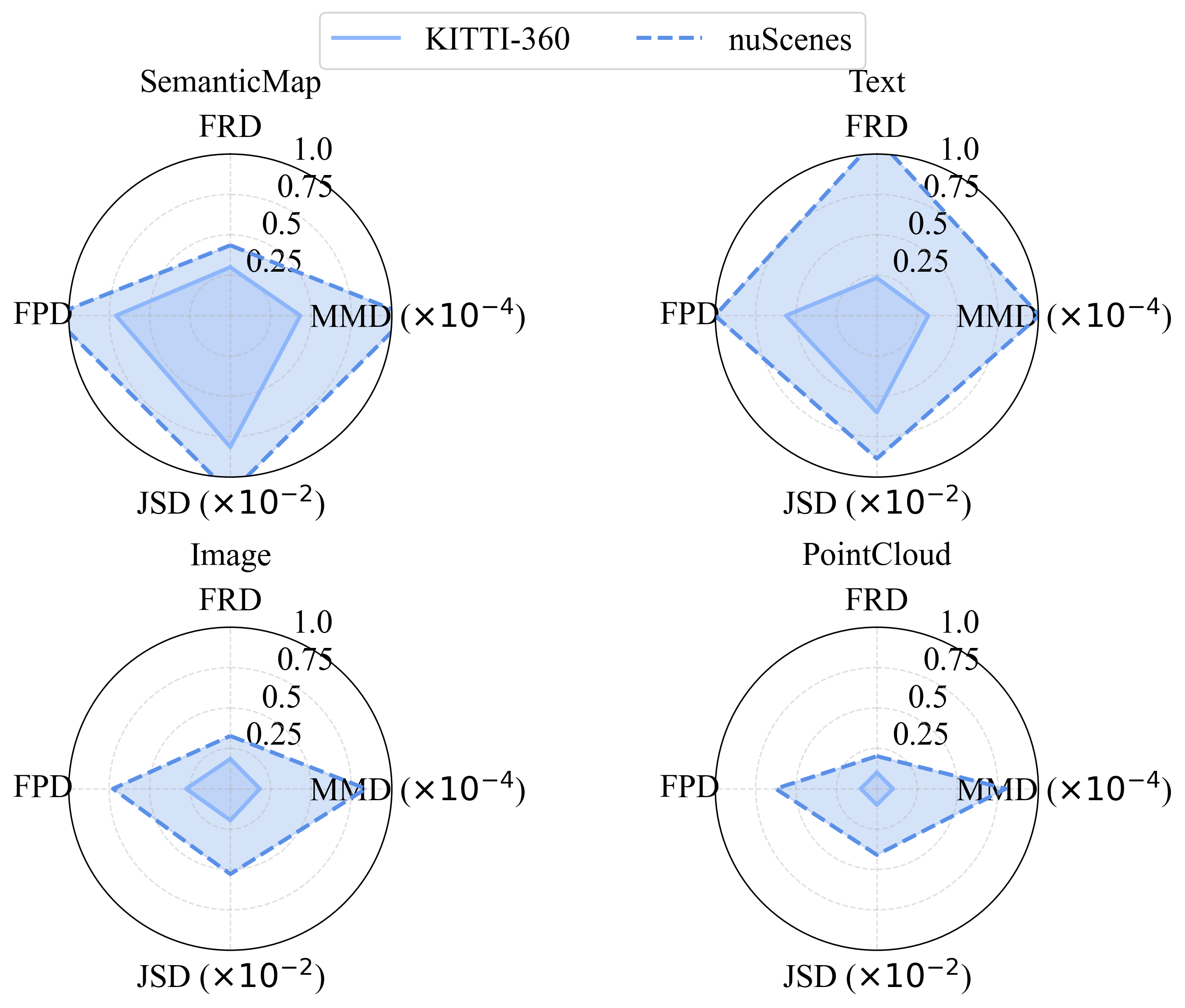}
\caption{A cross-dataset comparison shows that the model consistently achieves better performance on KITTI-360 than on nuScenes across all evaluation metrics.} 
\label{fig:nuscence}
\end{figure}

We train our model on the KITTI-360~\cite{kitti360} dataset, a large-scale suburban driving benchmark with panoramic imagery, dense LiDAR, and semantic labels. The model is trained on 69,580 samples from 8 sequences and evaluated on 11,517 samples. We first study how semantic conditioning affects geometric structure and object shape generation on KITTI-360. To assess generalization, we perform cross-dataset evaluation on KITTI-360 and nuScenes~\cite{nuscenes2019}, using layouts extracted from each dataset to compute metrics (\autoref{fig:nuscence}). We additionally test on SemanticKITTI~\cite{semantickitti2019} and present qualitative temporal results in \autoref{fig:semantickitti} . This multi-dataset evaluation demonstrates robustness across diverse environments, sensors, and traffic conditions.

\noindent\textbf{Implementation Details.} All point clouds were uniformly projected into 64×1024 range images, with training conducted on 4 NVIDIA RTX 3090 GPUs.
{ The training is split into two stages: we first train the unconditional generation model on the KITTI-360~\cite{kitti360} dataset, which serves as a foundation for conditional generation; we then freeze the parameters of the unconditional generation model and train the conditional generation model. We observed that freezing the ControlNet~\cite{controlnet} parameters while initially training only the zero convolution layers, followed by joint training of ControlNet and the zero convolution layers, leads to faster optimization This may be due to the strong alignment between our layout conditions and point cloud.}



\noindent\textbf{Experimental Setups.} We first assess our basic layout-based conditional LiDAR generation capabilities on the KITTI-360 dataset, and then evaluate the generation results based on several different modal conditions: image, text, and LiDAR point clouds. Finally, we conduct comprehensive ablation studies by removing or modifying key components of our framework and compare the results against state-of-the-art baselines,including~\cite{latent},~\cite{lidardiffusion},and~\cite{wu2024text2lidar}.

\noindent\textbf{Metrics.}We use the following metrics to evaluate the quality of generated point clouds, Frechet Range Distance(FRD), Minimum Matching Distance(MMD), Jensen-Shannon Divergence(JSD), Frechet Point Cloud and Distance(FPD). A comprehensive explanation of these evaluation metrics is provided in the supplementary material.

\subsection{Layout to LiDAR Point Cloud}\label{sec:layout2lidar}

\begin{figure}[t]
\centering
\includegraphics[width=0.5\textwidth]{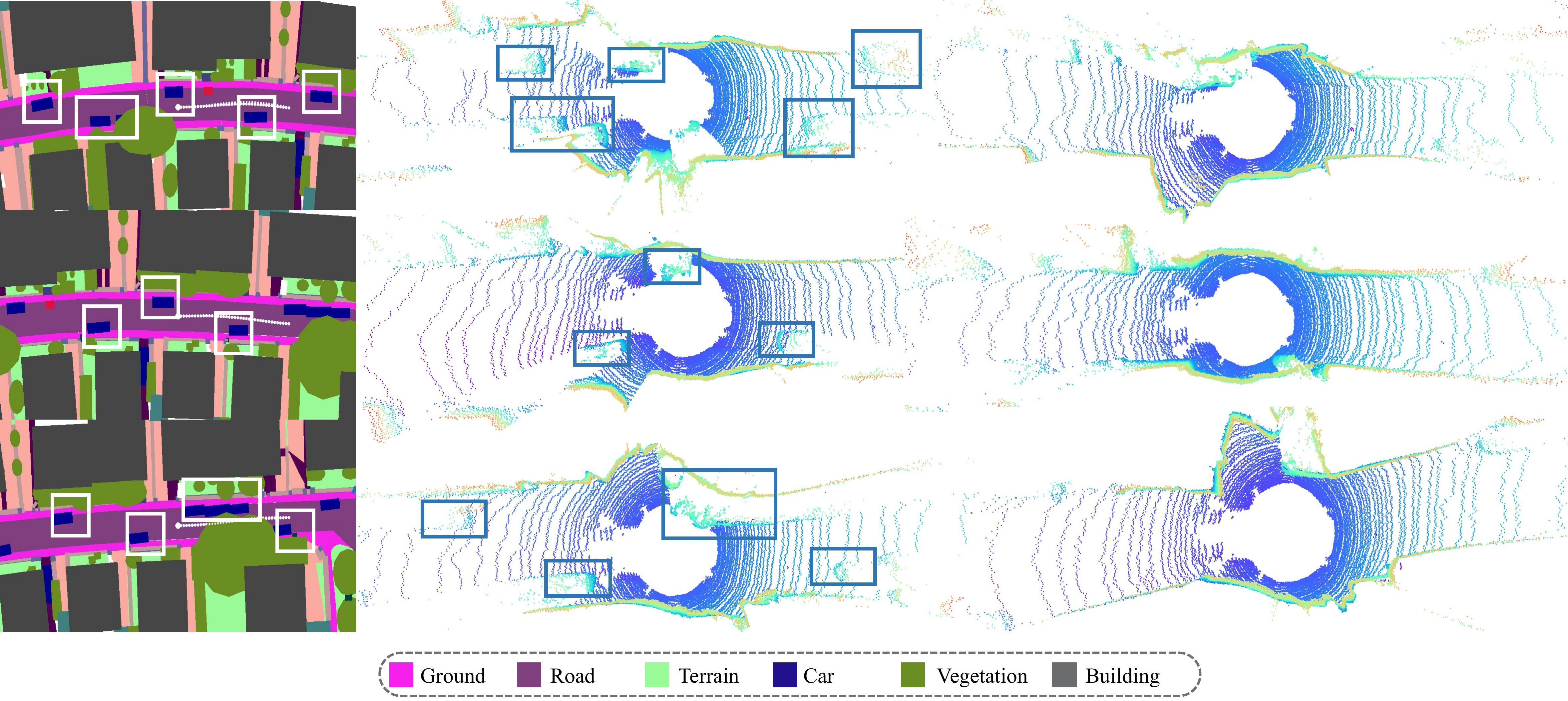}

\caption{\textbf{Layout to LiDAR Point Cloud results.} The white boxes indicate vehicle locations in the layout, and the blue boxes show the generated vehicle positions. The rightmost column presents samples of the remove-car manipulation.}
\label{fig:manlipulation}
\end{figure}




As previously mentioned, we trained a conditional generative model based on a unified layout representation. We then sample across various layouts and manipulate point clouds by modifying these layouts. \autoref{fig:manlipulation} illustrates our generation and manipulation results, showing that the input layouts effectively control the generated point clouds. The sampled point clouds display \textbf{smooth ground surfaces, straight road alignments, and accurately positioned vehicles with consistent shapes and sizes}.Furthermore, point cloud manipulations are highly efficient, as modifying the layouts alone allows seamless adjustments. We present the results after removing vehicles in \autoref{fig:manlipulation}.

\vspace{-6pt}
\begin{figure*}[t]
\includegraphics[width=1\textwidth]{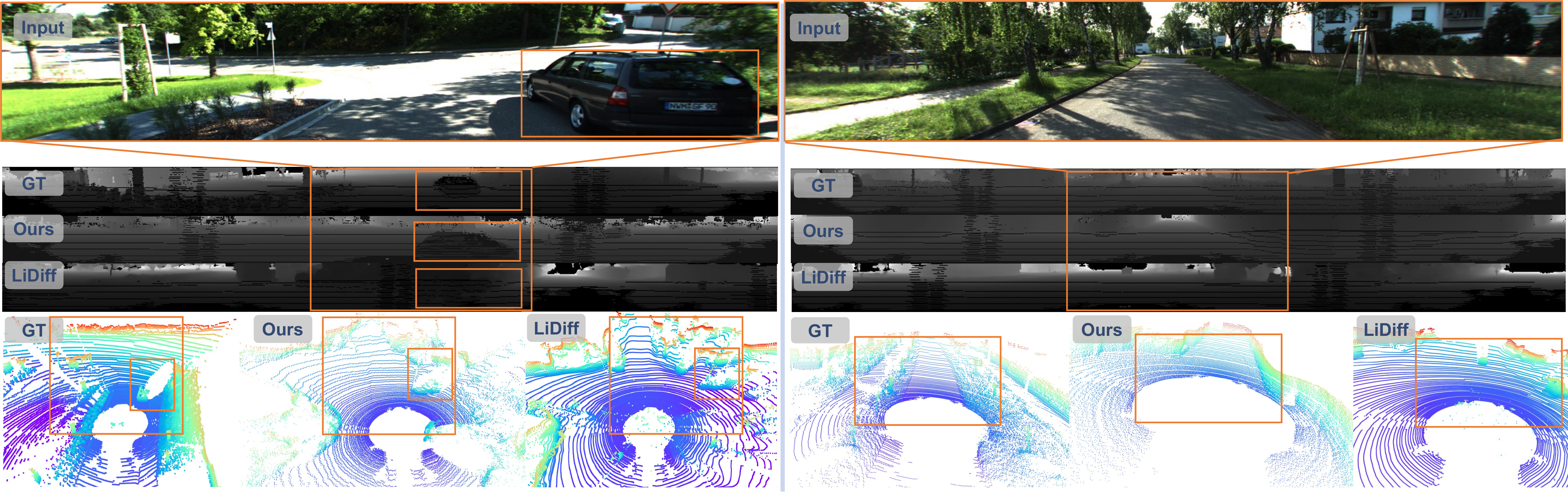}
\caption{\textbf{Image to LiDAR Point Cloud generation.} The orange boxes mark the area covered by the input image.In the left column, the vehicles in the input image are accurately sampled by both our method and LiDARDiffusion. However, the LiDARDiffusion sampling results contain multiple interfering vehicles. In the right column, the input image shows an empty road, and our method generates a point cloud along the straight path. However, LiDARDiffusion erroneously generates vehicles.}
\label{fig:image2pc}
\end{figure*}

\begin{figure*}[t]
\centering
\includegraphics[width=1\textwidth]{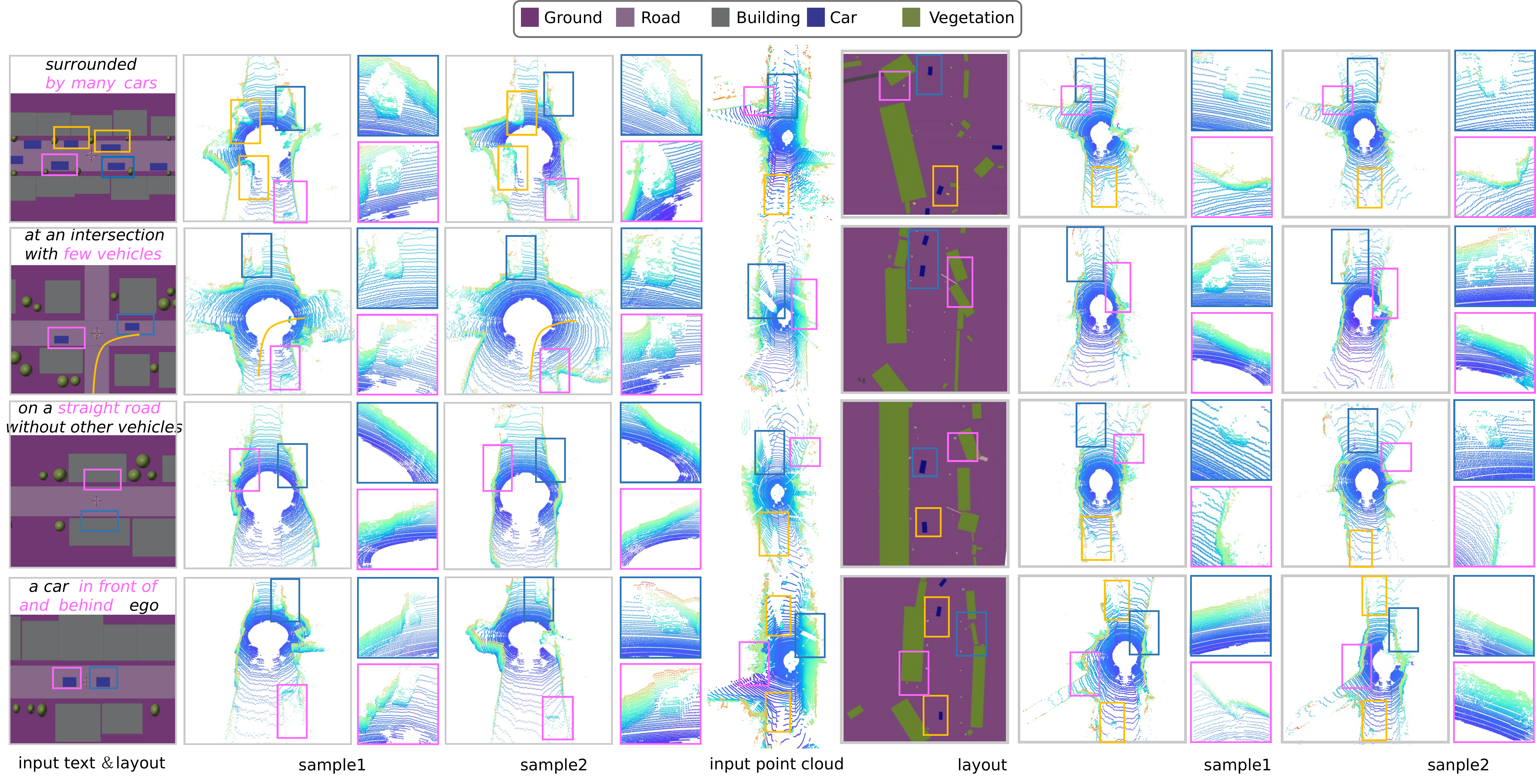}
\caption{\textbf{Text (left half) / Point Cloud (right half)  to LiDAR Point Cloud generation.} \textbf{For text}: an automated driving scene layout description is generated with GPT and used to sample the point cloud. LiDARDraft accurately samples the point cloud consistent with the text description, recognizing road types, vehicle counts, and their relative positions. The blue and pink boxes highlight specific details of the sampled point cloud. 
\textbf{For point cloud}: given a single-frame point cloud as input, LiDARDraft can sample multiple point clouds with consistent layouts and ensure layout consistency while maintaining diversity. The blue and pink boxes highlight the local details of the sampled point clouds.}
\label{fig:result}
\end{figure*}


\subsection{Image to LiDAR Point Cloud}\label{sec:image2lidar}
Cameras and LiDAR are currently the two most fundamental sensors in autonomous driving platforms. If generative models can bridge the gap between image and point cloud modalities, it will greatly advance driverless driving simulation.However, image-to-point cloud generation remains an under-explored task, with only LiDAR-Diffusion~\cite{lidardiffusion} making an attempt. Nevertheless, its performance is subpar, indicating the need for further research in this area. Due to this, we further explore the possibility of converting images into point clouds.Our method consists of three main steps: First, we employ the semantic segmentation model SAM~\cite{sam} and the depth estimation model DepthAnything~\cite{depth} to obtain the semantic and depth information for each pixel in the input image, which is then used to generate a pseudo point cloud. Second, we apply the DBSCAN clustering algorithm~\cite{DBSCAN} to the pseudo point cloud to derive the corresponding layout of the image. Finally, we sample the point cloud based on the generated layout.

As in \autoref{fig:image2pc}, our results significantly outperform the current state-of-the-art LiDARDiffusion. Since KITTI-360 only provides perspective images, we limit our comparison to the regions highlighted in orange. In the left column, both our method and LiDARDiffusion accurately sample the vehicles present in the input image. However, the results from LiDARDiffusion exhibit multiple spurious vehicles, introducing unintended artifacts. In the right column, where the input image depicts an empty road, our approach faithfully generates a point cloud aligned with the road structure. In contrast, LiDARDiffusion erroneously synthesizes non-existent vehicles, highlighting its limitations.

\subsection{Text to LiDAR Point Cloud}\label{sec:text2lidar}

Text to LiDAR Point Cloud becomes significantly more challenging in self-driving contexts, where the large-scale characteristics and scene complexity are challenging to describe with simple text. Recent advancements in large language models (LLMs)~\cite{gpt3,gpt5}  have started to address these limitations. In this work, we propose a novel approach that leverages GPT-5~\cite{gpt5} to generate layouts of automatic driving scenes.  
We input specific and precise scene descriptions into GPT-5, such as \textit{"a vehicle driving on a road with vehicles both in front and behind, with trees and houses along both sides of the road"}and request GPT-5 to generate the corresponding Blender code for the layout. The generated layout is then imported into Blender to create the 3D scene, from which we sample the corresponding LiDAR point cloud. \autoref{fig:result} shows our sampling results. It can be seen that LiDARDraft accurately identifies intersection types, assesses vehicle quantities, and recognizes spatial relationships.The arrangement of houses and trees along the road is neat, with no interference between objects. As illustrated on the left side of \autoref{fig:result}, we sampled multiple outputs that \textbf{not only align with the textual descriptions but also exhibit considerable diversity}. This capability offers a promising proof-of-concept for “from scratch simulation”, revealing the potential for directly generating autonomous driving environments using text prompts alone in the future.

\subsection{LiDAR Point Cloud Transformation}\label{sec:lidar2lidar}

LiDAR Point Cloud Transformation has significant applications across various fields. Studies such as Choi et al.\cite{choi2021part} and Hu et al.\cite{hu2023context} leveraged LiDAR Point Cloud Transformation for data augmentation, effectively enhancing the performance of 3D object detectors. We also apply LiDARDraft to point cloud transformation tasks. The right portion of \autoref{fig:result} further demonstrates  samples of the point cloud transformation results giving an input LiDAR-scanned point cloud. It shows that LiDARDraft samples point clouds that conform to the layout of the scan, and it maintains controlled variability.

This indicates that LiDARDraft supports the task of generating diverse point clouds with consistent layouts from a single-frame point cloud input and enables adaptive and realistic transformations between point clouds while preserving spatial characteristics. We can effectively utilize LiDARDraft to generate diverse samples to enhance robustness.


\begin{table*}[!htbp]
    \centering
    \footnotesize
    \caption{\textbf{Comparison of different condition types across various metrics.}}
    \begin{tabular}{lcccccccc}
        \toprule
        Metrics & FRD$\downarrow$ & MMD ($\times 10^{-4}$)$\downarrow$ & JSD$\downarrow$ & FPD$\downarrow$ 
                & FRD$\downarrow$ & MMD ($\times 10^{-4}$)$\downarrow$ & JSD$\downarrow$ & FPD$\downarrow$ \\
        \cmidrule(lr){1-9}
        Condition Type & \multicolumn{4}{c}{\textbf{Semantic Map}} & \multicolumn{4}{c}{\textbf{Text}} \\
        \cmidrule(lr){2-5} \cmidrule(lr){6-9}
        Latent Diffusion\cite{latent} & 24.21 & 3.31 & 0.088 & 20.60 & - & - & - & - \\
        LiDARDiffusion\cite{lidardiffusion} & 22.93 & 3.16 & 0.072 & 18.50 & 80.61 & 4.74 & 0.415 & 29.60 \\
        Text2LiDAR\cite{wu2024text2lidar} & - & - & - & - & 170.12 & 5.12 & 0.439 & 34.81 \\
        \rowcolor{blue!10}
        LiDARDraft(ours) & \textbf{21.72} & \textbf{3.14} & \textbf{0.070} & \textbf{18.32} & \textbf{23.88} & \textbf{3.23} & \textbf{0.079} & \textbf{19.11} \\
        \midrule
        Condition Type & \multicolumn{4}{c}{\textbf{Image}} & \multicolumn{4}{c}{\textbf{Point Cloud}} \\
        \cmidrule(lr){2-5} \cmidrule(lr){6-9}
        Latent Diffusion\cite{latent} & 50.50 & 3.99 & 0.314 & 26.51 & 129.90 & 4.99 & 0.434 & 32.98 \\
        LiDARDiffusion\cite{lidardiffusion} & 45.14 & 3.72 & 0.256 & 25.83 & 121.50 & 4.82 & 0.423 & 31.77 \\
        \rowcolor{blue!10}
        LiDARDraft(ours) & \textbf{24.62} & \textbf{3.39} & \textbf{0.094} & \textbf{21.32} & \textbf{27.98} & \textbf{3.43} & \textbf{0.099} & \textbf{22.12} \\
        \bottomrule
    \end{tabular}
    \label{tab:comparison}
\end{table*}

\begin{figure}
\centering
\includegraphics[width=0.5\textwidth]{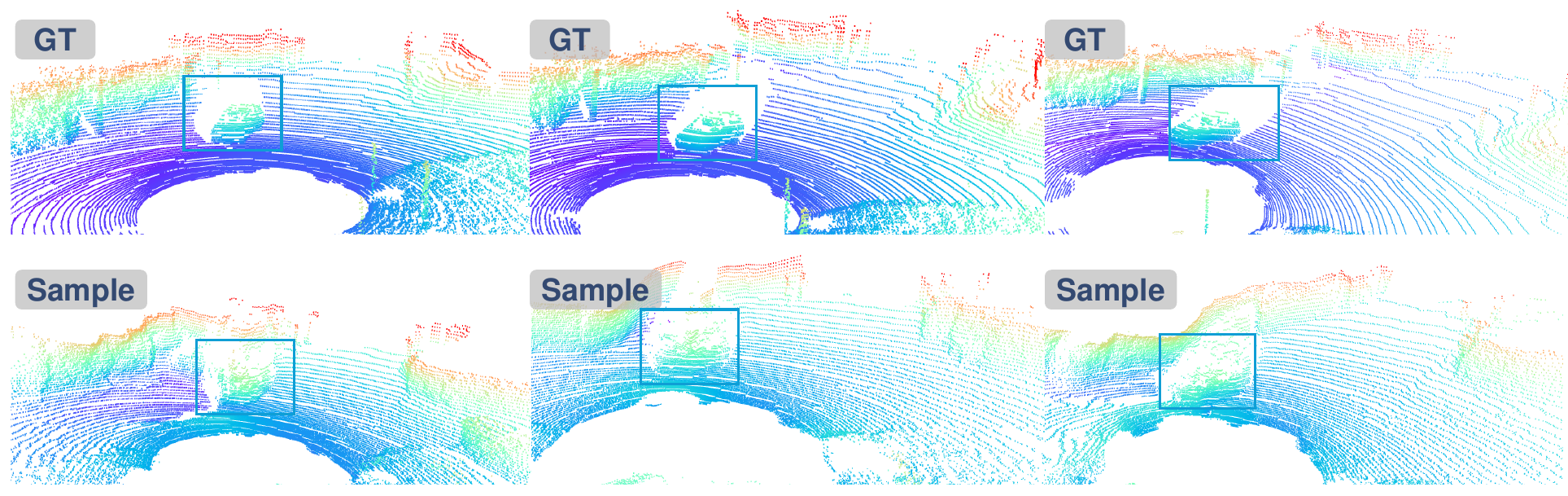}
\caption{Qualitative Results on the SemanticKITTI Dataset.
We select three consecutive LiDAR frames as the conditioning input and observe that the generated point clouds remain temporally consistent, geometrically plausible, and visually realistic across the sequence.} 
\label{fig:semantickitti}
\end{figure}

\vspace{-6pt}

\subsection{Efficiency Analysis}
\begin{figure}
\centering
\includegraphics[width=0.49\textwidth]{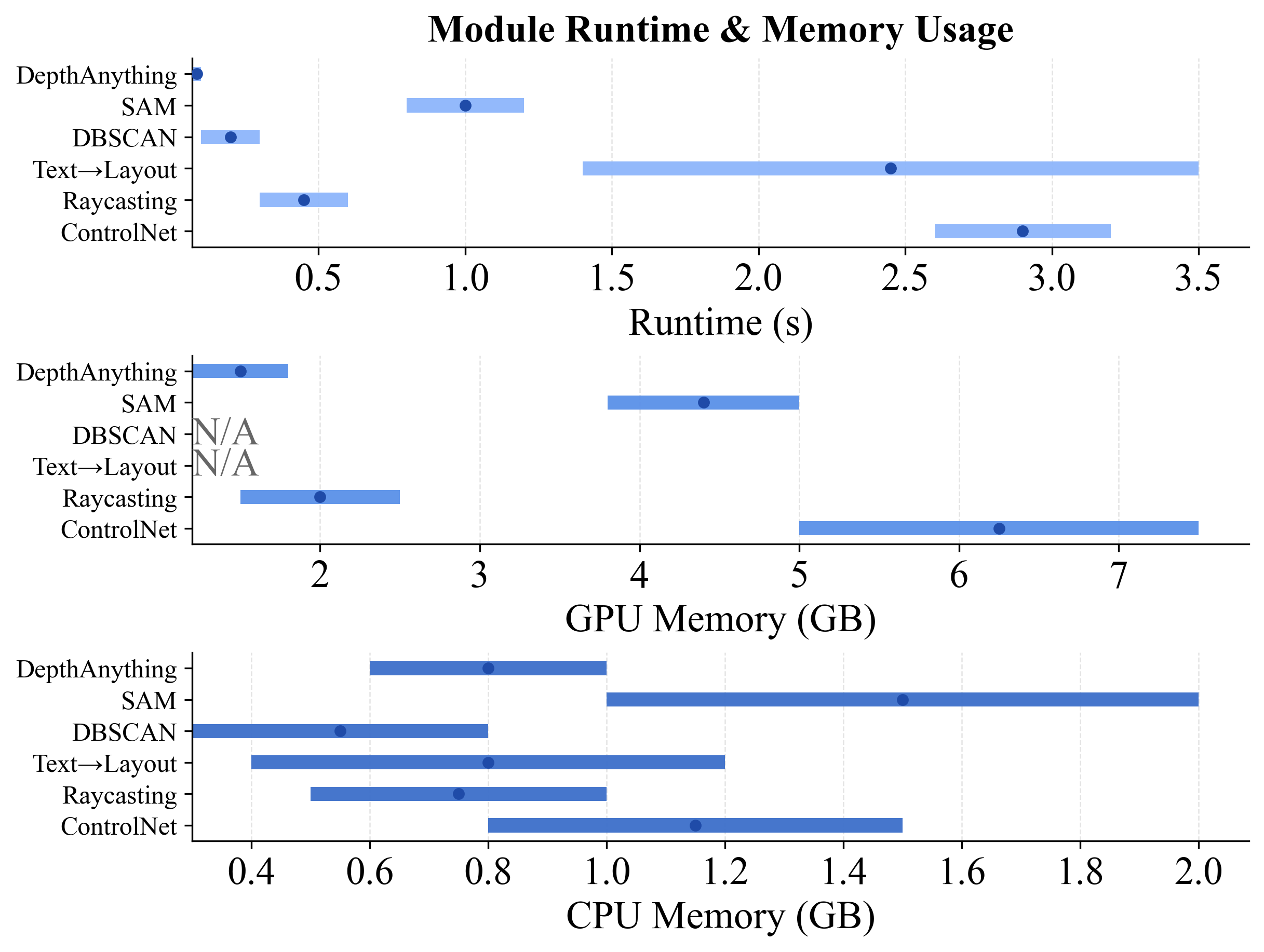}
\caption{Module Runtime and Memory Usage.} 
\label{fig:efficiency}
\end{figure}

We further examine the computational cost of each module in our pipeline using a single NVIDIA RTX 3090 GPU and an Intel-class CPU. As illustrated in \autoref{fig:efficiency}, the system exhibits a clear stratification of efficiency across perception, geometric processing, and generative components. The perception modules show a sharp contrast: lightweight depth estimation operates with minimal latency and memory footprint, while transformer-based semantic extraction requires noticeably more resources, forming the upper bound of the perception cost. CPU-bound geometric processing such as DBSCAN remains consistently efficient, with narrow runtime intervals and limited memory demand.

The Text→Layout module stands out with the widest runtime interval due to its CloudAPI-dependent execution, introducing external I/O overhead rather than computation-bound latency. In contrast, GPU-accelerated physical simulation (Raycasting) maintains stable and predictable cost regardless of scene variation. The generative module (ControlNet) forms the dominant GPU consumer, reflecting the inherent complexity of diffusion-based generation, yet its runtime remains bounded and does not dominate the overall pipeline.

Overall, the system displays a well-structured efficiency profile: lightweight perception and geometric modules form the majority, a single diffusion model accounts for the main GPU load, and a cloud-dependent module contributes the primary latency variability. This separation of computational regimes provides clear avenues for further acceleration, such as localizing text-to-layout generation or compressing the generative backbone.

\vspace{-3pt}

\begin{figure}
\centering
\includegraphics[width=0.5\textwidth]{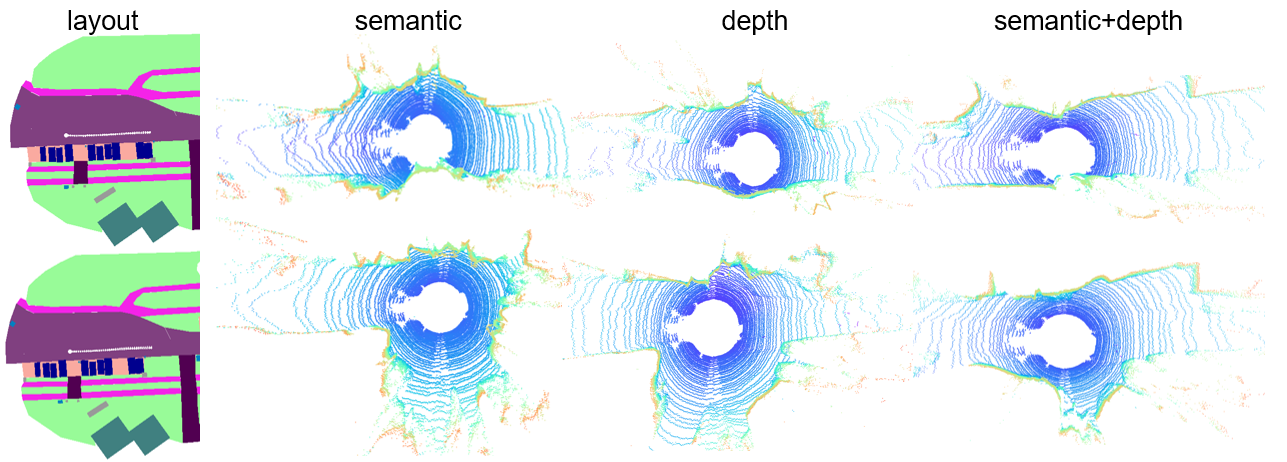}
\caption{Incorporating different channels leads to distinct effects on the sampled results: adding depth information enhances geometric fidelity, while incorporating semantic cues improves the structural accuracy of objects such as vehicles.} 
\label{fig:channel}
\end{figure}
\vspace{-3pt}


\subsection{Comparison with Baselines}
LiDARDraft unifies various conditional inputs, including semantic maps, text, images, and point clouds, into a shared layout representation to guide point cloud generation. As this is a novel and underexplored task, directly comparable baselines are limited. For each input modality, we carefully select the most relevant baselines for comparison.

Specifically, Latent Diffusion~\cite{latent} incorporates the four types of conditions into unconditional point cloud generation by concatenating them with the latent representation. Similarly, LiDARDiffusion~\cite{lidardiffusion}, originally supporting three input types, is modified to handle a fourth condition via the same concatenation operation. For text-based generation, we compare against Text2LiDAR~\cite{wu2024text2lidar}, which represents the current state-of-the-art.

Our method is the only one that supports all four conditional inputs. For semantic maps, layout control is achieved through ControlNet~\cite{controlnet}(see \autoref{fig:semantickitti}); for the other three inputs, our framework directly encodes them into layouts and generates samples accordingly. As shown in \autoref{tab:comparison}, our approach consistently achieves superior performance across all input types.

\subsection{Ablation Study}
We conduct ablation experiments to assess the contribution of each component (Table~\ref{tab:ablation}). Finetuning ControlNet proves essential—removing it leads to a substantial degradation, indicating that the model relies on this adaptation to capture scene-specific structures. Replacing our RayCast-based layout projection with a simpler BEV projection causes a pronounced drop across all metrics, showing the necessity of spatially accurate projection. Removing semantic segmentation further harms performance, demonstrating its role in preserving object-level structure, while omitting depth information produces a moderate decline, reflecting its contribution to geometric fidelity. Overall, the full model outperforms all ablated variants, verifying that each component provides complementary benefits for high-quality LiDAR point cloud synthesis.

\begin{table}
    \centering  
    \footnotesize
    \caption{\textbf{Ablation study on the key components of our method.}}
    \begin{tabular}{l@{\hspace{6pt}}c@{\hspace{4pt}}c@{\hspace{4pt}}c}
        \toprule
        Metrics & FRD $\downarrow$ & MMD ($\times 10^{-4}$) $\downarrow$ & JSD $\downarrow$ \\
        \midrule
        w/o finetune controlnet & 54.76 & 4.01 & 0.323 \\
        raycast $\rightarrow$ bev 
        & 123.66 & 4.91 & 0.429 \\
        w/o semantic segmentation & 28.98 & 3.67 & 0.122 \\
        w/o depth estimation    & 24.91 & 3.40 & 0.095 \\
        \rowcolor{blue!10}  
        ours                    & \textbf{23.01} & \textbf{3.19} & \textbf{0.077} \\
        \bottomrule
    \end{tabular}
    \label{tab:ablation}
\end{table}
\vspace{-6pt}


 
\section{Conclusion}
We present LiDARDraft, a novel conditional LiDAR point cloud generation framework that supports control inputs from text, image, and point cloud modalities. To our knowledge, LiDARDraft is the first approach to unify these three modalities, enabling seamless integration with unconditional generative models. LiDARDraft demonstrates that fully automated autonomous driving simulation, using textual descriptions, casually captured photos, or single-frame point cloud scans as inputs, is feasible in the future.




\FloatBarrier
\small
\bibliographystyle{ieeenat_fullname}
\bibliography{main}

\end{document}